\documentclass[10pt, a4paper]{article}

\usepackage[review]{lrec-coling2024} % this is the new style
\usepackage{float}
\title{NJUST-KMG at TRAC-2024 Tasks 1 and 2: Offline Harm Potential
Identification}

\name{Jingyuan Wang $^1$, Shengdong Xu $^1$, Yang Yang* $^1$}

\address{
\textsuperscript{1}Nanjing University of Science and Technology
\\
         Nanjing, Jiangsu, China \\
         \{WangJingyuan357, shengdong.xu, yyang\}@njust.edu.cn
         }

\abstract{
This report provide a detailed description of the method that we proposed in the TRAC-2024 Offline Harm Potential dentification which encloses two sub-tasks. The investigation utilized a rich dataset comprised of social media comments in several Indian languages, annotated with precision by expert judges to capture the nuanced implications for offline context harm. The objective assigned to the participants was to design algorithms capable of accurately assessing the likelihood of harm in given situations and identifying the most likely target(s) of offline harm. Our approach ranked second in two separate tracks, with F1 values of 0.73 and 0.96 respectively. Our method principally involved selecting pretrained models for finetuning, incorporating contrastive learning techniques, and culminating in an ensemble approach for the test set. 
 \\ \newline \Keywords{Social Media Analysis, Classification, Fine-tuning, Contrastive Learning} }

\begin{document}

\maketitleabstract

\section{Introduction}

The TRAC-2024 Offline Harm Potential Identification task is a critical effort aimed at addressing the pressing issues \citet{yang2023deep} regarding the impact of online content once taken into a real-world, offline context, broadly the task is to predict whether a specific post is likely to initiate, incite or further exaggerate an offline harm event (viz. riots, mob lynching, murder, rape, etc). With the exponential growth of digital platforms, monitoring the diverse and multilingual content becomes paramount to prevent detrimental consequences in social interactions and individual well-being. This task emphasizes the challenging aspect of understanding nuanced implications embedded within conversations in various Indian languages, highlighting the urgency in developing sophisticated models that can navigate the intricacies of linguistic and cultural nuances.  

Our system leverages the synergy of advanced pretrained models \citet{devlin2018bert} with the progressive concept of contrastive learning \citet{chen2020simple}, which have extensive applications in various fields such as multi-modal learning \citet{yang2019semi, yang2019semi_2}, continual learning, semi-supervised learning, etc. We harness the rich representations\citet{yang2024alignment} learned by models trained on extensive corpuses and tailor these to our specific context through meticulous fine-tuning. By integrating contrastive learning, we enhance the model's ability to discern subtleties within the dataset's multilingual content, crafting a more robust system against the diversity of languages and semantic complexities \citet{huo2018learning}. The ensemble strategy \citet{dietterich2000ensemble} employed at the testing phase not only solidifies the individual strengths of diverse models but also ensures our system's resilience and generalization across different data points.

Participating in the TRAC-2024 task offered profound insights into the content moderation and harm prediction landscape, especially concerning the subtleties involved in cross-linguistic and cultural contexts. The key decision to integrate contrastive learning into our methodology was driven by empirical observations during the development phase. Initial results indicated that our model exhibited difficulties in distinguishing between the top three categories of harm potential, often conflating instances with subtle differences. Recognizing the critical need for a clear delineation between these categories, we turned to contrastive learning as a strategic solution to enhance the discriminative capacity of our model. Contrastive learning, by design, operates on the principle of distinguishing between similar and dissimilar pairs of data, effectively 'pushing apart' representations of different categories while 'pulling together' representations of the same category. By implementing this approach, we aimed to increase the distance in the feature space between the harm potential categories, thereby reducing the ambiguity and improving the precision of our classifications. This methodological pivot was instrumental in addressing the nuances of multilingual content, which often requires a delicate balance of linguistic subtlety and cultural awareness to accurately identify and categorize harm potential indicators.

\section{Background}
\label{sec:append-how-prod}

The TRAC-2024 challenge comprised two sub-tasks designed to evaluate the offline harm potential of online content. As input, models received social media text data, extensively annotated to assess the harm potential, drawn from various Indian languages reflective of the region's diversity. In sub-task 1a, the output required was a four-tier classification that predicted the potential of a document to cause offline harm, ranging from 'harmless' to 'highly likely to incite harm.' An example input might be a social media post, and the output would be a categorical label from 0 to 3 indicating projected harm. In sub-task 1b, models predicted the potential target identities impacted by the harm, classifying them into categories such as gender, religion, and political ideology. Our participation focused on sub-task 1a, utilizing our expertise in dealing with subtle nuances of context and language.

Our approach took inspiration from existing research on using pretrained models for text classification \citet{yang2022domfn}, where methods like those described by \citet{devlin2018bert}. in the development of BERT have set foundations. What distinguishes our contribution is the incorporation of contrastive learning to refine these models within the multilingual context of Indian social media. This novel implementation aimed to enhance delineation among closely related content categories, addressing the challenge of high intra-class variation and inter-class similarity. Our method introduced an effective differentiation among content rated with varying levels of harm potential, thereby innovating within the established realm of text classification.

\section{Method}
\subsection{Base Model}
We adopted and compared several different pre-trained models, including XLM-R \citet{conneau2019unsupervised}, MuRILBERT \citet{khanuja2021muril} and Bangla-Bert \citet{sarker2022banglabert}, and some other models will be mentioned in subsequent experiments.
\subsubsection{XLM-R}
XLM-R is a transformer-based language model trained with the multilingual MLM objective on 100 languages \citet{razzak2019integrated}, two languages in the competition's dataset included. In order to deal with multi-language issues, XLM-R proposed new methods for data processing and model optimization objectives. The former uses Sentence Piece with a unigram language model to build a shared sub-word vocabulary, and the latter introduces a supervised optimization objective of translation language modeling(TLM). In this competition, we directly added a linear layer to fine-tune the pre-trained model for the classification tasks.

\subsubsection{MuRILBERT}

MuRIL (Multilingual Representations for Indian Languages) is a cutting-edge language model built on the transformer architecture, designed with the intention of enhancing natural language understanding for Indian languages. It provides superior performance over previous models by being pre-trained on a vast corpus covering 17 Indian languages, including transliterated text. MuRIL's innovation lies in its tailored pre-training regimen that caters to the nuanced syntactic and semantic structures unique to Indian languages, leveraging tasks like translated language modeling and transliteration invariance.
\subsubsection{BanglaBERT}
BanglaBERT, on the other hand, is a specialized transformer-based model meticulously honed for the Bengali language. It is pre-trained with a masked language model (MLM) objective on a large corpus of Bengali text sourced from diverse genres, ensuring a thorough representation of the language's contextual nuances. By adopting a language-specific approach, BanglaBERT presents a robust solution for various Bengali NLP tasks, encompassing both classical and advanced modeling techniques. In the context of this competition, akin to how XLM-R was adapted, we refined BanglaBERT with an additional linear layer, fine-tuning the pre-existing model to skillfully undertake classification challenges presented by the dataset.

\subsection{Strategy}
Our strategy is very simple: fine-tune the pre-trained model, adopt a comparative learning loss function, and finally perform model integration.
\subsubsection{Contrastive Learning}
Contrastive learning is a technique in machine learning that trains models to differentiate between dissimilar pairs of data while recognizing similarities among equivalent instances. This approach is particularly useful in settings where the objective is to learn accurate and distinct representations of data points that may otherwise appear to be closely related. 

At its core, contrastive learning utilizes pairs of data points, known as positive pairs, which are similar to each other, and negative pairs, which are not. Through various training strategies, a model is encouraged to output similar representations for positive pairs and distinct representations for negative pairs. This creates a more defined feature space, where the representations of different classes or categories are more separable, thus improving classification performance.The most commonly used loss is infoNCE, and the formula is as follows:
\begin{equation}L_{infoNCE}=-\frac1N\sum_{i=1}^N\log\frac{\exp(\sin(z_i,z_{i+}))/\tau)}{\sum_{j=1}^N\exp(\sin(z_i,z_{i,j}))/\tau)}\end{equation}
where $z_{i}$ is the feature representation of the i-th sample, $z_{i+}$ is its positive sample, and sim (x, y) is the similarity measure between samples x and y (for example cosine similarity), $\tau$ is a temperature parameter that controls the shape of the loss function.
\subsubsection{Model Ensemble}
Model ensemble is a sophisticated technique that amalgamates multiple distinct machine learning or deep learning models to substantially bolster the performance and stability of the overall predictive system. This approach capitalizes on the unique strengths of diverse models, integrating their predictions to mitigate individual biases and variances, thereby enhancing the ensemble's generalization capabilities. Utilizing methods such as voting, averaging, or stacking, the ensemble tailors its strategy to fit the problem at hand, adapting to various scenarios and maximizing advantages.

In the TRAC-2024 competition, we strategically deployed model ensemble to optimize accuracy, drawing upon an array of fine-tuned models imbued with insights specific to the complex, multilingual dataset at our disposal. Through selective aggregation of model predictions, our ensemble harnessed the collective intelligence of its constituents, effectively minimizing overfitting and capturing the essence of intricate linguistic nuances. The resultant system not only demonstrated superior performance but also maintained consistent reliability, validating the potency of model ensemble as a cornerstone of our methodology and a pivotal factor in achieving commendable F1 scores.

\section{Experiment}
\textbf{Dataset.} We treat sub-task a as a 4-class classification task, and sub-task b as a multi-label 5-classification \citet{yang2018complex, he2022not} task. Our approach is to fine-tune the pre-trained model using official datasets. But We split the training and validation sets randomly instead of following the official way, specifically, we divide the training set into two parts: training and validation, with a ratio of 4:1.\\
\textbf{Metric.} The evaluation metric for this competition is the F1 Score, which is the harmonic mean of Precision and Recall.\\
\textbf{Implementation Detail.} 
The maximum number of text tokens used by the language-model method is 512. Our approach is founded on fine-tune principles and makes use of the pre-trained model made available on the official xlm-roberta-base
\footnote{\label{myfootnote}$https://huggingface.co/xlm-roberta-base$}, xlm-roberta-large
\footnote{\label{myfootnote}$https://huggingface.co/FacebookAI/xlm-roberta-large$}, MuRILBERT\footnote{\label{myfootnote}$https://huggingface.co/google/muril-base-cased$},
BanglaBERT\footnote{\label{myfootnote}$https://huggingface.co/sagorsarker/bangla-bert-base$}. Regarding the hyperparameter settings of subtask b, we set the threshold $\eta =0.5$. We did the same data preprocessing as in the \citet{narayan2023hate}\\
\textbf{Comparison Methods Result.} It should be noted that in this competition we are mainly doing sub-task a, so the specific experimental results of sub-task a will be explained next. Table \ref{tab: compare} shows the F1 score performance, from which we can observe that: 1) There's a clear gradient in performance, with more sophisticated models generally achieving higher F1 scores. This suggests that models with greater complexity or those fine-tuned on domain-specific data tend to have better predictive capabilities for the given task. 2) IndicBERT \citet{kakwani2020indicnlpsuite} and BanglaHateBert \citet{jahan2022banglahatebert}, which likely are tailored to specific language datasets, perform less well compared to more general multilingual models. This could indicate that while language-specific models have an advantage in understanding linguistic nuances, they might lack the broader context that multilingual models are trained on. 3) There are two variants of the XLM-R model, base and large, both scoring the same F1 score of 0.70. This might imply that for this specific task, the additional parameters and complexity of the larger model do not add significant value over the base model. Alternatively, it could also indicate that the task is less sensitive to model size and more dependent on other factors such as dataset quality or training techniques. 4) The highest score is achieved by the Model Ensemble method (F1 score of 0.73), which outperforms the individual models. This exemplifies the main advantage of ensembles in integrating diverse predictive patterns, thereby improving generalizability and reducing errors that might be present in single models.\\
\textbf{Ablation Study.} To analyze the contribution of the contrastive loss and model ensemble strategy in our method, we conduct more ablation studies in this competition. The F1 score after adding contrastive loss is demonstrated in Table \ref{tab: abl1}. From the table, we can clearly see that after adding contrastive loss, the F1 value has a certain improvement. The f1 values under different model ensemble strategies are shown in the table \ref{tab: abl2}. It is obvious that the average ensemble method has the highest results.
\begin{table}[H]
\centering
\begin{tabular}{lll}
\hline
\textbf{Method} &  \textbf{F1}\\
\hline
\textbf{IndicBERT} & 0.44 \\
\textbf{BanglaHateBert} & 0.63 \\
$\textbf{Twitter-R}_{base}$ & 0.64 \\
\textbf{HateBERT} & 0.66 \\
\textbf{MuRILBERT} & 0.69 \\
\textbf{BanglaBERT} & 0.69 \\
$\textbf{XLM-R}_{base}$ & 0.70\\
$\textbf{XLM-R}_{large}$ & 0.70\\
\textbf{Model Ensemble}  & \textbf{0.73}\\
\hline
\end{tabular}
\caption{\label{tab: compare} Comparison method.}
\end{table}

\begin{table}[H]
\centering
\begin{tabular}{lll}
\hline
\textbf{Method} &  \textbf{F1}\\
\hline

\textbf{MuRILBERT} & 0.688 \\
$\textbf{MuRILBERT}_{contra}$ & \textbf{0.700} \\
\textbf{BanglaBERT} & 0.686 \\
$\textbf{BanglaBERT}_{contra}$ & \textbf{0.695} \\
\hline
\end{tabular}
\caption{\label{tab: abl1} Ablation experiment on contrastive loss.}
\end{table}

\begin{table}[H]
\centering
\begin{tabular}{lll}
\hline
\textbf{Method} &  \textbf{F1}\\
\hline

$\textbf{Model Ensemble}_{vote}$ & 0.723 \\
$\textbf{Model Ensemble}_{w-avg}$ & 0.730 \\
$\textbf{Model Ensemble}_{avg}$ & \textbf{0.731} \\
\hline
\end{tabular}
\caption{\label{tab: abl2} Ablation experiment on model ensemble strategies.}
\end{table}

\section{Conclusion}

Despite the strategic implementation of model ensemble techniques and contrastive learning in our approach for the TRAC-2024 competition, certain limitations were observed. The intricacy of the multilingual dataset and the subtlety of contextual nuances inherent in the social media comments called for an even finer granularity in modeling. Our ensemble, while robust, still faced challenges in dissenting rare language constructs and cultural idioms, which occasionally led to misclassifications. Moreover, the contrastive learning, albeit effective in distinguishing between categories with subtle differences, revealed a need for more sophisticated negative sampling strategies to fully capture the complex dynamics of potential offline harm in diverse cultural contexts. These shortcomings underscore areas for future research and refinement, in pursuit of a model with an even more nuanced understanding and predictive prowess.

\nocite{*}
\section{Bibliographical References}\label{sec:reference}

\bibliographystyle{lrec-coling2024-natbib}
\bibliography{lrec-coling2024-example}

\end{document}